\documentclass[]{interact}

\usepackage{epstopdf}
\usepackage[caption=false]{subfig}
\usepackage[onehalfspacing]{setspace}

\usepackage[numbers,sort&compress]{natbib}
\bibpunct[, ]{[}{]}{,}{n}{,}{,}
\makeatletter
\def\NAT@def@citea{\def\@citea{\NAT@separator}}
\makeatother

\usepackage{xcolor}
\usepackage{color,soul} 
\usepackage[%
unicode,
pdftex,
bookmarks=true,
bookmarksnumbered=true,
bookmarksopen=true,
bookmarksopenlevel=2,
colorlinks,
linkcolor=blue!80!black!80,
citecolor=blue!80!black!80,
linktocpage=false,
urlcolor=cyan,
]{hyperref}
\usepackage[open,openlevel=2]{bookmark}[2011/04/21]

\newcommand{\defeq}{\stackrel{\raisebox{-.1\height}{\tiny\rm def}}{=}}

\begin{document}


\title{Comment on ``All-optical machine learning using diffractive deep neural networks''}

\author{
\name{
 Haiqing Wei\textsuperscript{1},
 Gang Huang\textsuperscript{1},
 Xiuqing Wei\textsuperscript{2,*}\thanks{$^*$\!~Corresponding authors: Xiuqing Wei (weixq@mail.sysu.edu.cn), Yanlong Sun (ysun@tamhsc.edu), Hongbin Wang (hwang@tamhsc.edu)},
 Yanlong Sun\textsuperscript{3,*},
 Hongbin Wang\textsuperscript{3,*}
}
\affil{\textsuperscript{1}Ambow Research Institute, Ambow Education Group, Beijing 100088, China; \textsuperscript{2}Department of Gastroenterology, The Third Affiliated Hospital of Sun Yat-sen University, Guangzhou 510630, China; \textsuperscript{3}Center for Biomedical Informatics, College of Medicine, Texas A\&M University, Houston, TX, USA}
}

\maketitle

\begin{abstract}
Lin {\it et al.} (Reports, 7 September 2018, p. 1004) reported a remarkable proposal that employs a passive, strictly linear optical setup to perform pattern classifications. But interpreting the multilayer diffractive setup as a deep neural network and advocating it as an all-optical deep learning framework are not well justified and represent a mischaracterization of the system by overlooking its defining characteristics of perfect linearity and strict passivity.
\end{abstract}

Lin {\it et al.} \cite{Lin18} proposed a combination of methods for creating a computer-generated volumetric hologram (CGVH) made of multiple planar diffractive elements, and using such hologram to scatter and directionally focus each of a multitude of pattern-imprinted coherent light fields into a designated spatial region on an image sensor, effectively realizing a functionality of pattern recognition and classification. Their all-optical multi-planed setup bears a certain resemblance to the multi-layered structure of a deep neural network (DNN) \cite{Goodfellow16}, but that is about as far as the similarity goes.

It is a mischaracterization to interpret the CGVH construct as a DNN, when its functionality is strictly limited to linear transformations of the input light field, thus unable to perform any task of statistical inference/prediction beyond the capacity of a single layer perceptron \cite{Goodfellow16,MinskyPapert1969}. Apart from the glaring absence of nonlinear activations therein, the passive CGVH setup also lacks parameter tunability to support neural network learning, except for, perhaps, the spacings between diffractive elements.

As such, the authors' claim of their CGVH providing ``{\em an all-optical deep learning framework in which the neural network is physically formed by multiple layers of diffractive surfaces}'' is perilously confusing and misleading, by overly stressing the superficial similarity between a multi-planed optical diffractive setup and a multilayer neural network, and glossing over a wide range of technical challenges in implementing a truly all-optical machine learning mechanism. A CGVH optical network remains strictly linear, where the linearity severely limits the achievable computations but lends convenience and mathematical rigor to analyses on the possible functionalities and performance limitations, while a multilayer neural network requires some nonlinearity to prevent layer collapsing, where the quintessential nonlinearity routinely defies rigorous mathematical analysis but affords Turing-complete computations.

To a broader audience who are familiar with linear optics, Lin {\it et al.}'s CGVH setup and working principles are reminiscent of volume hologram \cite{Goodman05}, volume optics \cite{Piestun10}, and optical mode converters \cite{Miller12,Morizur10,Labroille14}. Therefore, the richly developed linear theory of wave optics and mechanics can be applied to analyze the performance of a CGVH-based pattern classifier, using a rigorous theory of light propagation and scattering \cite{Goodman05}. Duly noting the all-important linearity is not merely a scholastic preference or a rhetorical option. Rather, it has significant theoretical and practical ramifications. It would be inexcusable and a disservice to willfully neglect the vast literature and results on the physics and mathematics of linear systems.

Without loss of generality and not underestimating its information and communication capacities, a CGVH optical network can be considered as consisting of $L{+}2$ layers of planar diffractive elements, with $L\in\mathbb{Z}$, $L\ge 0$, and each panel of diffractive elements contains no more than $N{\times}N$ resolvable pixels to transmit, receive, or modulate a light field, such that each panel is completely characterized by an $N^2$-dimensional complex-valued vector. The $0$-th layer under a coherent illumination generates an input of amplitude image field represented by a $N^2$-dimensional complex-valued vector $\psi_0(j)$, $j\in[1,N^2]$, while the $(L{+}1)$-th layer is an image detector that does no better than reporting a $N^2$-dimensional real-valued vector of optical intensities $|\psi_{L{+}1}(j)|^2$, $j\in[1,N^2]$, subject to unavoidable noise. For each $l\in[1,L]$, the $l$-th panel of diffractive elements is characterized by a diagonal matrix $Q_l={\rm diag}(\{e^{-a_{jl}+ib_{jl}}\})$, with $a_{jl},\,b_{jl}\in\mathbb{R}$ representing the absorption and phase delay by the $j$-th pixel, $\forall j\in[1,N^2]$.

The free-space wave propagation between the $(l{-}1)$-th and the $l$-th planes, $\forall l\in[1,L{+}1]$, can be described by a linear operator $P_l={\cal F}^{-1}{\rm diag}(\{e^{i\beta(k)D_l}\}){\cal F}$, with ${\cal F}$ denoting the unitary matrix of Fourier transform which turns a real space image $\{\psi_{l{-}1}(j)\}$ into a spatial frequency image $\{{\cal F}[\psi_{l{-}1}](k)\}$, $k\in\mathbb{Z}$ indexing a plane wave with an associated phase velocity $\beta(k)$ along the optical axis normal to the planes, and $D_l$ is the spacing between the two planes in question. Naturally, ${\cal F}^{-1}$ represents the inverse Fourier transform. Then, $\forall l\in[1,L]$, the amplitude field incident on the $l$-th panel becomes $P_l\psi_{l{-}1}$, which is then pixel-wise modulated by the $l$-th panel of diffractive elements and turned into $\psi_l\defeq Q_lP_l\psi_{l{-}1}$. The cascade of optical diffractions continues until an amplitude field $\psi_L\defeq\left(\Pi_{l=1}^LQ_lP_l\right)\!\psi_0$ exits from the $L$-th and last panel of diffractive elements, which finally propagates to the detector and becomes $\psi_{L{+}1}=M\psi_0$, with $M\defeq P_{L{+}1}\!\left(\Pi_{l=1}^LQ_lP_l\right)$.

A crucially important fact is that the entire optical setup can be described by a single $N^2{\times}N^2$ matrix $M$, regardless of $L$, even if it approaches infinity and the setup becomes a continuous volume hologram. $M$ is a contraction operator, with all of its singular values upper bounded by $1$. The truth of the matter is that, not only the linear transformations of any series of passive optical elements give rise to a single contraction operator $M$, but also any given contraction operator $M$ as a wave field transformation can be realized through a serious of free-space propagations and point-wise amplitude modulations \cite{Borevich81}. Moreover, any linear optical device can be considered as an optical mode converter \cite{Miller12}. Linearity is key to the possibility of lumping multiple steps of operations into a single matrix of transformation and the amenability to rigorous mathematical analyses using matrix algebra. It is regrettable that reference \cite{Lin18} failed to seize upon and exploit the opportunity. One of the most important consequences of strict linearity in a CGVH-based all-optical pattern classifier is that the pattern discrimination power (PDP) becomes severely limited, being no better than the Euclidean distance discriminator, as will be proven rigorously below via an inequality on vector norms.

Consider two differently patterned input images $\psi_0$ and $\phi_0$ producing amplitude fields $\psi_{L{+}1}=M\psi_0$ and $\phi_{L{+}1}=M\phi_0$ on the detector plane, which square into intensity images $\Psi_{L{+}1}\defeq|\psi_{L{+}1}|^2$ and $\Phi_{L{+}1}\defeq|\phi_{L{+}1}|^2$, and induce proportional electric signals with the addition of unavoidable noise. In Lin {\it et al.}'s proposal, each intensity image is projected into a $K$-dimensional vector, $K\in\mathbb{N}$, by partially integrating the image within each of $K$ designated spatial areas. The PDP of the all-optical pattern classifier derives from a difference between $K$-dimensional vectors due to different patterns, which is well characterized and upper-bounded by the {\em total variation distance} (TVD) $\left\|\Psi_{L{+}1}-\Phi_{L{+}1}\right\|_1\defeq\sum_{j,p}\left|\psi_{L{+}1}^2(j,p)-\phi_{L{+}1}^2(j,p)\right|$, with $p\in\{\mathrm{real},\mathrm{imag}\}$ indexing the real or imaginary part of a complex value. The TVD is in turn upper-bounded as
\begin{align}\label{DiscrimSignalBound}
\begin{split}
   \left\|\Psi_{L{+}1}-\Phi_{L{+}1}\right\|_1
   &= {\textstyle{\sum_{j,p}}}\left|\psi_{L{+}1}(j,p)-\phi_{L{+}1}(j,p)\right| \left|\psi_{L{+}1}(j,p)+\phi_{L{+}1}(j,p)\right| \\
   &\leq \left\|\psi_{L{+}1}-\phi_{L{+}1}\right\|_2 \left\|\psi_{L{+}1}+\phi_{L{+}1}\right\|_2 \\
   &= \left\|M(\psi_0-\phi_0)\right\|_2 \left\|M(\psi_0+\phi_0)\right\|_2 \\
   &\leq \left\|\psi_0-\phi_0\right\|_2 \left\|\psi_0+\phi_0\right\|_2 \\
   &\leq \left\|\psi_0-\phi_0\right\|_2 \left(\|\psi_0\|_2+\|\phi_0\|_2\right) ,
\end{split}
\end{align}
where $\|\!\cdot\!\|_2$ denotes the $L^2$ norm, corresponding to the total light power of an optical image. With image fields normalized, $\|\psi_0\|_2=\|\phi_0\|_2=1$, the TVD is upper-bounded by $2\left\|\psi_0-\phi_0\right\|_2$. Therefore, the PDP of an all-optical pattern classifier, devoid of any nonlinear activation that resembles a biological or artificial neuron, does not go beyond the classical Euclidean distance algorithms \cite{Michie94,Bishop06}. When two different images are in close similarity that the $L^2$ distance $\left\|\psi_0-\phi_0\right\|_2$ is small and below a certain noise level, the system of Lin {\it et al.}'s will have a hard time to tell $\psi_0$ and $\phi_0$ apart, no matter how obviously different they are to human eyes, or how easily they may be distinguished by a bona fide deep neural network with nonlinear activation functions.

By contrast, a canonical DNN has at least one hidden layer that implements a nonlinear activation function \cite{Goodfellow16}. A true optical DNN would require nonlinear optical interactions, although nonlinearity purely in the optical domain is notoriously hard to induce. Indeed, such difficulty of all-optical nonlinearity has prompted proposals and demonstrations \cite{Shen17,Chang18} of optical-electronic hybrid networks, or nonlinear networks incorporating optoelectronic devices that involve photoabsorption or photoelectric generation of charge carriers. Nonlinearity could potentially enable a DNN to escape the TVD bound of PDP in (\ref{DiscrimSignalBound}). While deep learning in the presence of nonlinear activation functions, or fundamentally the input-output behavior of a typical nonlinear network, is not as well understood mathematically as a linear system, it is generally believed that nonlinearity endows a DNN with certain computational power for better performances in learning and predicting, specifically, noise suppression and pattern discrimination. Distributed nonlinear activations in a DNN have the potential to regulate signals and suppress noise and detrimental inferences, much like distributed signal regeneration in a long-haul communications network \cite{Leclerc02,Simon03}. In other fields such as quantum computing, it is known that weak and distributed nonlinear amplitude evolution could amplify a small difference between initial states into drastically different and easily distinguishable output results \cite{Abrams98}.

In closing, despite the shortcomings, Lin {\it et al.}'s report still represents a significant contribution to interdisciplinary researches at the intersections of many scientific and technological fields, including volume optics, linear transformations, 3D printing, and of course pattern recognition/classification as the authors proposed originally. The reported results of numerical simulations and experimental tests have indicated potentials of coherent light diffraction through a volume hologram to serve as a linear classifier for pattern recognition applications. The optical multi-planed setup demonstrated an efficient implementation of a complicated linear transformation, namely, a sophisticated optical mode converter. Following Lin {\it et al.}'s trail blazing, the interested scientific and engineering community will no doubt get busy and start working on both the theoretical fundamentals and the engineering practicals. But the first thing in business is to place the CGVH setup in the right technical context, recognize its characteristic linearity, and take advantage of a vast literature and an immense knowledge accumulation in the related fields.

\section*{Disclosure statement}
The authors have no potential financial or non-financial conflicts of interest.

\section*{Notes on contributors}
All authors contributed equally in researching, collating, and writing.

\bibliographystyle{tfnlm}

\end{document}